\documentclass{article}

\PassOptionsToPackage{numbers, compress}{natbib}
\usepackage[final]{nips_2018}

\usepackage[utf8]{inputenc} % allow utf-8 input
\usepackage[T1]{fontenc}    % use 8-bit T1 fonts
\usepackage{hyperref}       % hyperlinks
\usepackage{url}            % simple URL typesetting
\usepackage{amsfonts}       % blackboard math symbols
\usepackage{nicefrac}       % compact symbols for 1/2, etc.
\usepackage{microtype}      % microtypography

\usepackage{graphicx}
\usepackage[font=small]{caption}
\usepackage{subcaption}
\usepackage{adjustbox}
\usepackage{wrapfig}
\graphicspath{{figures/pdf/}{figures/png/}}
\DeclareGraphicsExtensions{.pdf,.png}

\usepackage{array}
\usepackage{booktabs}
\usepackage{makecell}
\usepackage{tabu}
\usepackage{wrapfig}

\usepackage{algorithm}
\usepackage{algorithmic}

\usepackage{tikz}
\usetikzlibrary{arrows}
\usetikzlibrary{backgrounds}

\usepackage{xcolor}
\usepackage{calc} 
\usepackage{enumitem}

\usepackage{macro}

\title{On the Complexity of Exploration in Goal-Driven Navigation}

\author{
    Maruan~Al-Shedivat$^1$\thanks{Equal contribution. Correspondence: \texttt{\{alshedivat,lslee\}@cs.cmu.edu}.} \\
    \And
    Lisa~Lee$^{1*}$ \\
    \And
    Ruslan~Salakhutdinov$^{1,2}$ \\
    \And
    Eric~P.~Xing$^{1,3}$ \\
    \AND
    $^1$\textnormal{Carnegie Mellon University}, $^2$\textnormal{Apple}, $^3$\textnormal{Petuum Inc.}
}

\begin{document}

\maketitle

%!TEX root = ../main.tex

\begin{abstract}
Building agents that can explore their environments intelligently is a challenging open problem.
In this paper, we make a step towards understanding how a hierarchical design of the agent's policy can affect its exploration capabilities.
First, we design \texttt{EscapeRoom} environments, where the agent must figure out how to navigate to the exit by accomplishing a number of intermediate tasks (\emph{subgoals}), such as finding keys or opening doors.
Our environments are procedurally generated and vary in complexity, which can be controlled by the number of subgoals and relationships between them.
Next, we propose to measure the complexity of each environment by constructing dependency graphs between the goals and analytically computing \emph{hitting times} of a random walk in the graph.
We empirically evaluate Proximal Policy Optimization (PPO) with sparse and shaped rewards, a variation of policy sketches, and a hierarchical version of PPO (called HiPPO) akin to h-DQN.
We show that analytically estimated \emph{hitting time} in goal dependency graphs is an informative metric of the environment complexity.
We conjecture that the result should hold for environments other than navigation.
Finally, we show that solving environments beyond certain level of complexity requires hierarchical approaches.
\end{abstract}

%!TEX root = ../main.tex

\section{Introduction}\label{sec:introduction}

Deep reinforcement learning research has led us to discover general-purpose algorithms for learning how to control robots~\citep{levine2016end} and solve games~\citep{mnih2015human,silver2017mastering}, surpassing human abilities.
These results indicate a significant progress in the field.
However, building agents capable of intelligent exploration even in simple environments is still an unreached milestone.

To make progress towards this goal, first we need to understand and be able to measure when exploration is necessary.
For instance, while Atari games seem like a challenging benchmark, it turns out that having a memoryless reactive policy is often sufficient for solving most of these games~\citep{dann2016memory}.
On the other hand, there are environments (\eg, Montezuma's Revenge) that can only be solved by achieving some intermediate goals (\emph{subgoals}).
Learning about the dependencies between the subgoals requires executing a consistent exploration strategy, reasoning, and multi-step planning, beyond vanilla deep RL methods.

Broadly, exploration is a mechanism used by an agent to reduce uncertainty about its environment (\ie, rewards and state transitions).
Notable approaches to exploration include: (1) \emph{count-based} and \emph{intrinsic motivation} methods \citep{bellemare2016unifying}, where the agent (approximately) quantifies uncertainty of the states and actions and tends to visit the states it is least certain about; and (2) various policy-perturbation heuristics, such as $\varepsilon$-greedy, Boltzmann, and parameter-noise methods \citep{fortunato2017noisy,plappert2017parameter}.
All these approaches function on the level of atomic actions and hence are limited when it comes to complex structured tasks with delayed and sparse rewards.
To overcome such limitations, it is possible to use the framework of temporal abstractions (\emph{options}) \citep{sutton1999between,bacon2017option}.
In particular, \citet{kulkarni2016hierarchical} argued for hierarchical methods that enable exploration in the \emph{space of goals}, which is also our focus.

% Our work: measure of the environment complexity, 
In this paper, we aim to understand and measure the complexity of exploration in environments with multiple dependent subgoals, and the effects of hierarchical design of the agent's policy in such environments.
To do so, we introduce a collection of procedurally generated, simple grid-world environments called \texttt{EscapeRooms} (Table~\ref{tab:env-escaperoom}).
We represent the goal space with dependency graphs, and propose to measure complexity of exploration as the time it takes a random walk in this abstract space to reach the final goal state from the start state in expectation (\ie, the \emph{expected hitting time}). This measure captures a simple intuition: the more complex the goal dependencies are, the more time it would take the agent to explore how to solve the environment.

To verify that the hitting time is a useful measure of complexity in RL scenarios, we train a few hierarchical and non-hierarchical policies using methods based on proximal policy optimization \citep[PPO,][]{schulman2017proximal} on \texttt{EscapeRooms} and measure their exploration capabilities.
We use metrics such as success rate and the number of timesteps it takes the agent to achieve each goal.
Our results demonstrate that information about the goals is crucial to enable learning in our environments.
Moreover, we show that our complexity measure correlates with the performance of the policies---agents perform worse and hierarchy becomes more important in environments with higher exploration complexity.

%!TEX root = ../main.tex

\section{Methods}\label{sec:methods}

Given an environment, we would like to quantify \emph{how much exploration is needed to solve the task}.
In this section, we introduce the notion of goal-dependency graphs, describe \texttt{EscapeRoom} environments, and compute different measures of the exploration complexity for these environments.

\subsection{Goal-dependency graphs \& exploration complexity}

We are interested in a scenario where the agent can achieve the final goal only after having accomplished a number of intermediate goals.
Assuming that the goals and dependencies are given, we can construct a graph $G(V, E)$ with nodes $V$ representing the goals and edges $E$ representing the relationships between the goals.
From this perspective, we can treat the agent executing a (stochastic) policy in an environment with these subgoals as a random walk on the corresponding goal-dependency graph.
To measure complexity of exploration in the given environment, we introduce the following notation.
Let $n$ be the number of nodes in the graph, and $W \in \Rb^{n \times n}$ the adjacency matrix of the graph weighted by the probabilities of transition from one goal to the other (according to the policy $\pi$ executed by the agent).
Let $D \in \Rb^{n \times n}$ be the corresponding diagonal weighted degree matrix:
\begin{equation*}
    D_{ii} := \sum_{j=1}^n W_{ij}, \quad D_{ij} := 0,\, \forall i \neq j
\end{equation*}
Now, we can use the graph Laplacian, $L := W - D$, to compute the expected time it would take the random walk to reach a given goal node with index $t$ from the initial node with index $s$ in the graph for the first time (also known as the \emph{hitting time}) \citep{lovasz1993random}.
To do so, we can solve the following linear system (subscripts denote indices):
\begin{align}
%\begin{split}
    L x &= b \quad \mathrm{s.t.}\ x_t = 0, \label{eq:hitting-time-laplacian} \\%\quad
    \mathrm{where} \quad b_s &= 1,\, b_t = -1,\, b_k = 0 \quad \forall k \notin \{s, t\} \nonumber
%\end{split}
\end{align}
where $x, b \in \Rb^n$.
The solution, $x^\star_s$, will be the hitting time from $s$ to $t$.
Solving \eqref{eq:hitting-time-laplacian} for each goal in the graph allows us to analytically compute different statistics for any goal-dependency graph under a given random walk, \eg, the expected number of states reachable under a given time limit.

\subsection{\texttt{EscapeRoom} environments}

We design a set of grid-world environments (Table~\ref{tab:env-escaperoom}) where the agent must pick up keys and open locked doors (\ie, accomplish intermediate goals) in order to reach the exit (the final goal).
The agent has 5 actions: \texttt{move-forward}, \texttt{turn-left}, \texttt{turn-right}, \texttt{pick-up} (key), and \texttt{open} (door).
The \texttt{pick-up} action only succeeds if the key is in front of the agent.
The \texttt{open} action only succeeds if a locked door is in front of the agent and the agent has already picked up a key of the same color as the door.
Upon arriving at the exit, the agent receives a reward of 1 and the episode terminates. 
Our \texttt{EscapeRoom} environments are based on Gym MiniGrid~\cite{gym_minigrid} and follow the OpenAI Gym API~\citep{brockman2016openai}.

\def\varoffset{0.50}
\def\varlabel{-0.35}
\def\varstart{0}
\def\varone{\varstart+\varoffset}
\def\vartwo{\varstart+2*\varoffset}
\def\varthree{\varstart+3*\varoffset}
\def\varfour{\varstart+4*\varoffset}
\def\varfive{\varstart+5*\varoffset}
\def\varsix{\varstart+6*\varoffset}
\def\varseven{\varstart+7*\varoffset}

\def\imagewidth{68pt}

\begin{wraptable}{r}{6.7cm}
    \begin{minipage}[b]{6.7cm}
    \caption{%
    \texttt{EscapeRoom} environments.
    On the right, we enumerate all possible dependency graphs (up to a permutation of colors) for environments with two rooms (a), three rooms (b, c), and four rooms (d, e, f ,g).
    Each node in the dependency graph can be traversed at most once (\ie, no cyclic paths are allowed).
    The agent (\textcolor{nice_red}{red triangle}) must pick up keys and open locked doors in order to reach the exit (\textcolor{nice_green}{green square}).
    Each door can only be opened by a key of the corresponding color.
    }\label{tab:env-escaperoom}
    \vspace{-1ex}
    \tabulinesep =_4pt^4pt
    \begin{tabu}to \columnwidth{@{}X[-2.5,l,m]X[-2.5,r,m]@{}}
        \tabucline[1pt]{-}
        Goal dependency graphs & Environment \\ \tabucline[0.6pt]{-}
        \begin{minipage}{\imagewidth}
        \begin{tikzpicture}[->, >=stealth', shorten >=1pt, auto, node distance=1cm, thick, main/.style={circle,draw}, x=1.4cm, y=3cm]
            \def\ymid{1}
            \node[draw=none] (label2) at (\varlabel, \ymid) {(a)};
            \node[main, fill=gray!30, draw=black!50, label={above:\scriptsize{start}}] (room0) at (\varstart, \ymid) {};
            \node[fill=nice_red!75, draw=black!50, label={above:\scriptsize{key}}] (key1) at (\varone, \ymid) {};
            \node[main, fill=nice_red!75, draw=black!50, label={above:\scriptsize{door}}] (room1) at (\vartwo, \ymid) {};
            \node[main, fill=green!75, draw=black!50,label={above:\scriptsize{exit}}] (exit) at (\varthree, \ymid) {};
            \draw (room0) to (key1);
            \draw (key1) to (room1);
            \draw (room1) to (exit);
        \end{tikzpicture}
        \end{minipage} &
        \begin{minipage}{\imagewidth}
        \includegraphics[width=\imagewidth]{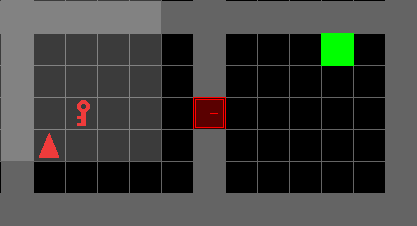}
        % \vspace{-15pt}
        \begin{center}
            \scriptsize{Sample environment (a)}
        \end{center}
        % \vspace{-5pt}
        \end{minipage} \\ \tabucline[0.6pt off 2pt]{-}
        \begin{minipage}{\imagewidth}
        \vspace{25pt}
        \begin{tikzpicture}[->, >=stealth', shorten >=1pt, auto, node distance=1cm, thick, main/.style={circle,draw}, x=1.5cm, y=3cm]
            % Escape room 1-1-1
            \def\ymid{1.4}
            \node[draw=none] (label2) at (\varlabel, \ymid) {(b)};
            \node[main, fill=gray!30, draw=black!50,label={above:\scriptsize{start}}] (room20) at (\varstart, \ymid) {};
            \node[fill=blue!75, draw=black!50] (key21) at (\varone, \ymid-0.1) {};
            \node[fill=purple!75, draw=black!50,label={above:\scriptsize{key}}] (key22) at (\varone, \ymid+0.1) {};
            \node[main, fill=blue!75, draw=black!50] (room21) at (\vartwo, \ymid-0.1) {};
            \node[main, fill=purple!75, draw=black!50,label={above:\scriptsize{door}}] (room22) at (\vartwo, \ymid+0.1) {};
            \node[main, fill=green!75, draw=black!50,label={above:\scriptsize{exit}}] (exit2) at (\varthree, \ymid-0.1) {};
            \draw (room20) to (key21);
            \draw (key21) to (room21);
            \draw (room20) to (key22);
            \draw (key22) to (room22);
            \draw (room21) to (exit2);
            \draw (room22) to (key21);

            % Escape room 2-0-0
            \def\ymid{0.9}
            \node[draw=none] (label2) at (\varlabel, \ymid) {(c)};
            \node[main, fill=gray!30, draw=black!50] (room10) at (\varstart, \ymid) {};
            \node[fill=blue!75, draw=black!50] (key11) at (\varone, \ymid) {};
            \node[main, fill=blue!75, draw=black!50] (room11) at (\vartwo, \ymid) {};
            \node[fill=purple!75, draw=black!50] (key12) at (\varthree, \ymid) {};
            \node[main, fill=purple!75, draw=black!50] (room12) at (\varfour, \ymid) {};
            \node[main, fill=green!75, draw=black!50] (exit1) at (\varfive, \ymid) {};
            \draw (room10) to (key11);
            \draw (key11) to (room11);
            \draw (room11) to (key12);
            \draw (key12) to (room12);
            \draw (room12) to (exit1);
        \end{tikzpicture}
        \end{minipage} &
        \begin{minipage}{\imagewidth}
        \vspace{-18pt}
        \includegraphics[width=\imagewidth]{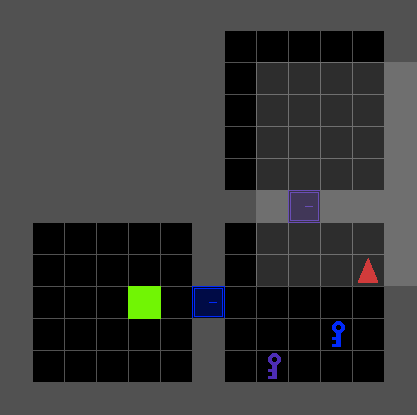}
        \vspace{-15pt}
        \begin{center}
            \scriptsize{Sample environment (b)}
        \end{center}
        \vspace{-5pt}
        \end{minipage} \\ \tabucline[0.6pt off 2pt]{-}
        \begin{minipage}{\imagewidth}
        \vspace{5pt}
        \begin{tikzpicture}[->, >=stealth', shorten >=1pt, auto, node distance=1cm, thick, main/.style={circle,draw}, x=1.5cm, y=3cm]
            % Escape room 3-0-0
            \def\ymid{0}
            \node[draw=none] (label1) at (\varlabel, \ymid) {(g)};
            \node[main, fill=gray!30, draw=black!50] (room10) at (\varstart, \ymid) {};
            \node[fill=blue!75, draw=black!50] (key11) at (\varone, \ymid) {};
            \node[main, fill=blue!75, draw=black!50] (room11) at (\vartwo, \ymid) {};
            \node[fill=nice_red!75, draw=black!50] (key12) at (\varthree, \ymid) {};
            \node[main, fill=nice_red!75, draw=black!50] (room12) at (\varfour, \ymid) {};
            \node[fill=purple!75, draw=black!50] (key13) at (\varfive, \ymid) {};
            \node[main, fill=purple!75, draw=black!50] (room13) at (\varsix, \ymid) {};
            \node[main, fill=green!75, draw=black!50] (exit1) at (\varseven, \ymid) {};
            \draw (room10) to (key11);
            \draw (key11) to (room11);
            \draw (room11) to (key12);
            \draw (key12) to (room12);
            \draw (room12) to (key13);
            \draw (key13) to (room13);
            \draw (room13) to (exit1);

            % Escape room 1-1-2
            \def\ymid{0.35}
            \node[draw=none] (label2) at (\varlabel, \ymid) {(f)};
            \node[main, fill=gray!30, draw=black!50] (room20) at (\varstart, \ymid) {};
            \node[fill=nice_red!75, draw=black!50] (key21) at (\varone, \ymid-0.1) {};
            \node[fill=blue!75, draw=black!50] (key22) at (\varone, \ymid+0.1) {};
            \node[main, fill=nice_red!75, draw=black!50] (room21) at (\vartwo, \ymid-0.1) {};
            \node[main, fill=blue!75, draw=black!50] (room22) at (\vartwo, \ymid+0.1) {};
            \node[fill=purple!75, draw=black!50] (key23) at (\varthree, \ymid-0.1) {};
            \node[main, fill=purple!75, draw=black!50] (room23) at (\varfour, \ymid-0.1) {};
            \node[main, fill=green!75, draw=black!50] (exit2) at (\varfive, \ymid-0.1) {};
            \draw (room20) to (key21);
            \draw (key21) to (room21);
            \draw (room20) to (key22);
            \draw (key22) to (room22);
            \draw (room21) to (key23);
            \draw (key23) to (room23);
            \draw (room23) to (exit2);
            \draw (room22) to (key21);
            
            % Escape room 1-1-2
            \def\ymid{1.5}
            \node[draw=none] (label3) at (\varlabel, \ymid) {(d)};
            \node[main, fill=gray!30, draw=black!50,label={above:\scriptsize{start}}] (room30) at (\varstart, \ymid) {};
            \node[fill=nice_red!75, draw=black!50] (key31) at (\varone, \ymid-0.1) {};
            \node[fill=blue!75, draw=black!50,label={above:\scriptsize{key}}] (key32) at (\varone, \ymid+0.1) {};
            \node[main, fill=nice_red!75, draw=black!50] (room31) at (\vartwo, \ymid-0.1) {};
            \node[main, fill=blue!75, draw=black!50,label={above:\scriptsize{door}}] (room32) at (\vartwo, \ymid+0.1) {};
            \node[fill=purple!75, draw=black!50] (key33) at (\varthree, \ymid-0.1) {};
            \node[main, fill=purple!75, draw=black!50] (room33) at (\varfour, \ymid-0.1) {};
            \node[main, fill=green!75, draw=black!50,label={above:\scriptsize{exit}}] (exit3) at (\varthree, \ymid+0.1) {};
            \draw (room30) to (key31);
            \draw (key31) to (room31);
            \draw (room30) to (key32);
            \draw (key32) to (room32);
            \draw (room31) to (key33);
            \draw (key33) to (room33);
            \draw (room32) to (exit3);
            \draw (room31) to (key32);
            \draw (room33) to (key32);

            % Escape room 1-1-2
            \def\ymid{0.9}
            \node[draw=none] (label4) at (\varlabel, \ymid) {(e)};
            \node[main, fill=gray!30, draw=black!50] (room30) at (\varstart, \ymid) {};
            \node[fill=nice_red!75, draw=black!50] (key31) at (\varone, \ymid) {};
            \node[fill=blue!75, draw=black!50] (key32) at (\varone, \ymid+0.2) {};
            \node[main, fill=nice_red!75, draw=black!50] (room31) at (\vartwo, \ymid) {};
            \node[main, fill=blue!75, draw=black!50] (room32) at (\vartwo, \ymid+0.2) {};
            \node[fill=purple!75, draw=black!50] (key33) at (\varone, \ymid-0.2) {};
            \node[main, fill=purple!75, draw=black!50] (room33) at (\vartwo, \ymid-0.2) {};
            \node[main, fill=green!75, draw=black!50] (exit3) at (\varthree, \ymid-0.2) {};
            \draw (room30) to (key31);
            \draw (key31) to (room31);
            \draw (room30) to (key32);
            \draw (key32) to (room32);
            \draw (room30) to (key33);
            \draw (key33) to (room33);
            \draw (room33) to (exit3);
            \draw (room31) to (key32);
            \draw (room31) to (key33);
            \draw (room32) to (key33);
            \draw (room32) to (key31);
        \end{tikzpicture}
        \end{minipage} &
        \begin{minipage}{\imagewidth}
        \vspace{-20pt}
        \includegraphics[width=\imagewidth]{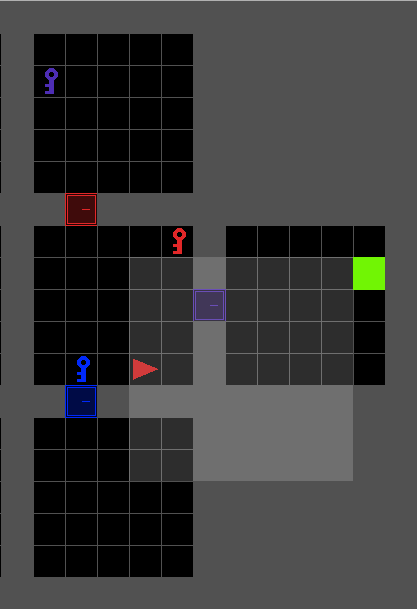}
        \begin{center}
            \scriptsize{Sample environment (f)}
        \end{center}
        \end{minipage} \\ \tabucline[1pt]{-}
    \end{tabu}
    \end{minipage}
\end{wraptable}

In each episode, we procedurally generate a new environment; the object locations, colors of the keys and doors, and the room layouts are all randomized.
The agent always begins at a random cell in the center room, which branches out to 1-3 other rooms, one of which contains the exit.
Each of the branching rooms is initially blocked by a locked door, so an environment with $n$ rooms has exactly $n-1$ keys and doors.
Each open cell can contain up to one of 3 object types (\texttt{Exit}, \texttt{Key}, or \texttt{Door}) with one of 6 possible colors.
The environment is partially observable, meaning that the agent can only observe its local surroundings and cannot see through walls. In our experiments, each observation is a $7 \times 7 \times 3$ array representing the $7 \times 7$ view in front of the agent with three channels (object IDs, color IDs, and a binary matrix capturing whether a door is open).

\textbf{Goal dependencies in \texttt{EscapeRooms}.}
In Table~\ref{tab:env-escaperoom}, we enumerate all possible goal dependency graphs for different \texttt{EscapeRoom} environments with up to 4 rooms.
Each goal is represented as one-hot encodings of (color, object); for example, (yellow key) means to pick up the yellow key, and (blue door) means to open the blue door.
To understand how complex each environment is from the stand point of exploration in the goal space, we compute the hitting time (HT) for each graph (Table~\ref{tab:laplacian}).
Note that dependency graphs in Table~\ref{tab:env-escaperoom} are simplified for illustration purposes (each goal node is assumed to be visited only once).
Computing properties of random walks requires strongly connected graphs, and hence we construct augmented goal-dependency graphs and use those for estimating the hitting times of interest (see Appendix~\ref{app:laplacian}).

\textbf{Complexity of \texttt{EscapeRooms}.}
Based on Table~\ref{tab:laplacian}, we make a few observations.
First, longer paths from the start to the exit nodes result in slower discovery of how to solve the environment.
Similarly, adding alternative paths that do not lead to the exit (proportional to the graph width) also increase the complexity and is reflected by the hitting time metric.
Finally, the environment complexity depends not only on the spatial map design but also on the action space.
We experimented with adding an extra \texttt{drop} (key) action which significantly increased the hitting time for the exit node in goal-dependency graph (Table~\ref{tab:laplacian}, last row).

\subsection{RL algorithms}\label{sec:rl-algs}

In this work, we focus on 
a class of policy gradient methods known as 
Proximal Policy Optimization (PPO) algorithms~\cite{schulman2017proximal}.
First, we evaluate the vanilla PPO trained using two different reward functions: (1) \textbf{PPO} is trained using sparse rewards, where the agent receives +1 reward upon achieving the final goal (\eg, reaching the exit); and (2) \textbf{PPO+Bonus}
is trained using reward shaping where in addition to the reward for achieving the final goal, the agent also receives +1 reward for achieving intermediate goals (\eg, picking up keys or opening doors). 

\newpage

    \begin{wrapfigure}{r}{0.48\textwidth}
    \vspace{-12pt}
    \begin{minipage}{0.48\textwidth}
    \begin{algorithm}[H]
        \caption{Hierarchical PPO}
        \label{alg:hippo}
        \begin{algorithmic}[1]
           \STATE {\bfseries Input:} Meta-controller policy $\pi_M$,\\
           Controller policy $\pi_C$
           \FOR{$i=1$ {\bfseries to} num\_episodes}
            \STATE subgoal $g \gets \pi_M(s)$
            \WHILE{$s$ is not terminal}
                \STATE $F \gets 0$
                \STATE $s_0 \gets s$
                \WHILE{{\bfseries not} ($g$ is reached)}
                    \STATE $a \gets \pi_C(\{s, g\})$
                    \STATE state $s'$, reward $f \gets \textrm{Env}(a)$
                    \STATE intrinsic reward $r \gets \textrm{Critic}(s', g)$
                    \STATE PPO\_update$(\pi_C, s, a, s', r)$
                    \STATE $F \gets F + f$
                    \STATE $s \gets s'$
                \ENDWHILE
                \STATE PPO\_update$(\pi_M, s_0, g, s', F)$
                \STATE subgoal $g \gets \pi_M(s)$
            \ENDWHILE
           \ENDFOR
        \end{algorithmic}
    \end{algorithm}
    \end{minipage}
    \vspace{-10pt}
    \end{wrapfigure}

Next, we introduce a variant of PPO called \textbf{HiPPO (Hierarchical PPO)} which borrows the hierarchical framework from~\cite{kulkarni2016hierarchical}, but replaces the hierarchical value functions in the h-DQN with hierarchical PPO policies. In more detail, HiPPO uses a \emph{meta-controller} policy to choose intermediate goals for the lower-level \emph{controller} policy to achieve\footnote{The action space of the meta-controller is the space of goals. The controller uses available primitive actions.}.
The controller receives one-hot encoded goals as part of its observation and \emph{intrinsic} rewards for achieving intermediate goals chosen by the meta-controller.
The meta-controller receives sparse \emph{extrinsic} rewards from the environment for achieving the final goal and is prompted to submit a new action (\ie, a new goal) each time the lower-level controller accomplishes the previous goal.
The pseudocode for HiPPO is given in Algorithm~\ref{alg:hippo}. In our experiments, we used a fixed meta-controller that chooses a sequence of goals along a random depth-first search path on the goal dependency graph, rather than a trainable meta-controller policy.

Lastly, \textbf{PPO+Sketch} is a variation of policy sketches~\citep{andreas2016modular} where the agent is provided with a sequence of goals that leads to achieving the final goal.
PPO+Sketch is identical to PPO except that in each timestep, the current observation is concatenated with the current intermediate goal\footnote{Feeding goals as observations into the policy network is slightly different from the original design of \citet{andreas2016modular}. We plan to investigate the original policy sketch architecture in future work.}, \ie, the actions produced by the policy are always conditional on the current goal.
Similar to PPO, and unlike HiPPO and PPO+Bonus, PPO+Sketch does not use intrinsic rewards for achieving intermediate goals.

%!TEX root = ../main.tex

\section{Experiments}
\label{sec:experiments}

We evaluate PPO, PPO+Bonus, PPO+Sketch, and HiPPO on \texttt{EscapeRoom} environments (a)-(g). We limit the episode length to 1000 time steps.
For each method and environment, we use the LSTM policy with hidden dimension 64, and train for 10M total time steps on 128 vectorized environments using the Adam optimizer, learning rate 2.5e-4, discount factor $\gamma=0.9$, and TD $\lambda=0.95$. We evaluated each method and environment over 5 trials with different random seeds.

\begin{figure*}[t]
    \centering
    \includegraphics[width=\textwidth]{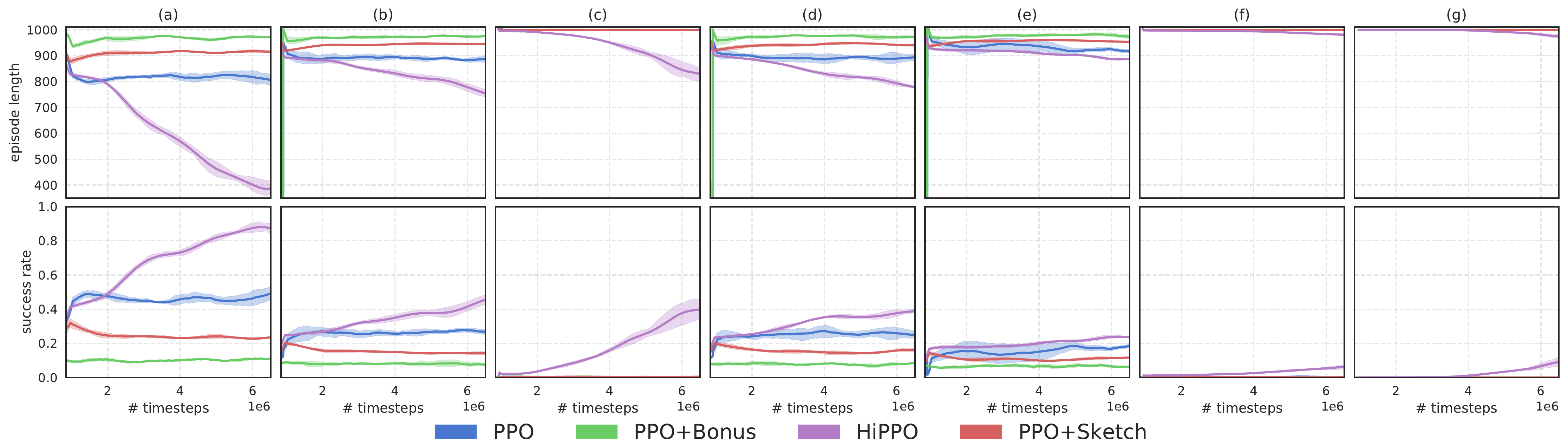}
    \caption{Average episode length and success rate on \texttt{EscapeRoom} environments with goal dependency graphs (a)-(g) from Table~\ref{tab:env-escaperoom}. In all environments, HiPPO achieves the best performance (smallest episode length and highest success rate). In the most complex environments (f) and (g), HiPPO still makes some learning progress.}
    \label{fig:learning-plot}
\end{figure*}

\begin{figure*}[h!]
    \centering
    \includegraphics[width=\textwidth]{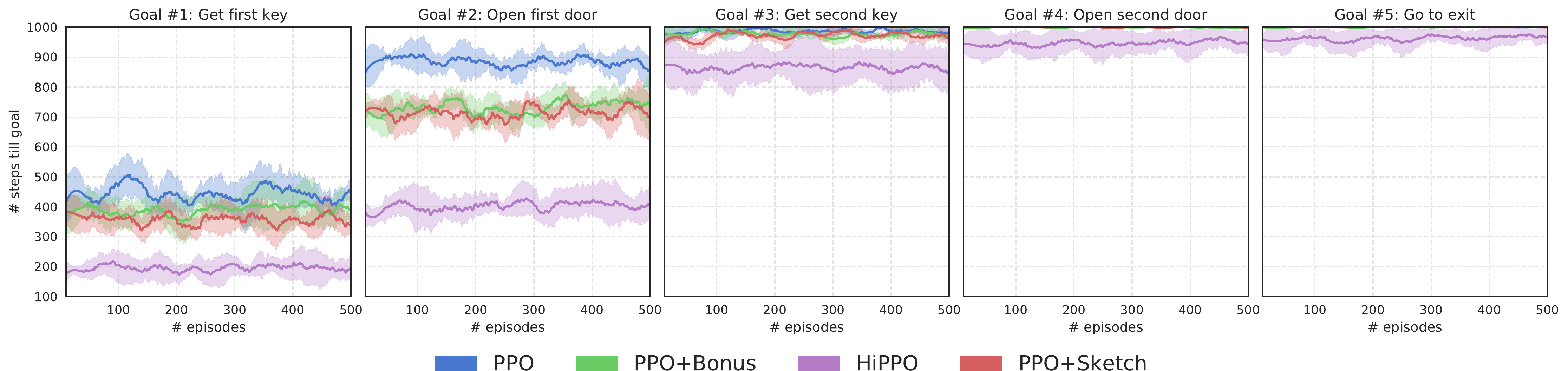}
    \caption{Average number of timesteps to reach each intermediate goal on \texttt{EscapeRoom} (c). HiPPO is the quickest method to achieve each goal.}
    \label{fig:goal-progress}
\end{figure*}

In Figures~\ref{fig:learning-plot} and \ref{fig:goal-progress}, we see that HiPPO consistently achieves the smallest average episode length and highest success rate on all environments, thus demonstrating the benefit of using hierarchical policies that operate at different temporal scales.
Surprisingly, PPO with sparse rewards performs better than PPO+Bonus, showing that the bonus rewards for achieving intermediate goals does not help a non-hierarchical policy.
We also find that PPO+Sketch performs worse than PPO indicating that merely conditioning on subgoals might be suboptimal and destructively interferes with optimization.

    \begin{wraptable}{r}{0.5\textwidth}
        \vspace{-10pt}
        \caption{%
        Depth, width, and hitting time (HT) statistics computed for \texttt{EscapeRoom} environments (a)-(g).}
        \label{tab:laplacian}
        \vspace{-1ex}
        \scriptsize
        \setlength{\tabcolsep}{4pt}
        \begin{tabular}{@{}lrrrrrrr@{}}
        \toprule
                            & (a) & (b) & (c) & (d) & (e) & (f) & (g) \\
        \midrule
        exit depth          &  2  &  2  &  4  &  2  &  2  &  4  &  6  \\
        graph width         &  1  &  2  &  1  &  2  &  3  &  2  &  1  \\
        HT (w/o drop-key)   &  8.4 & 12.1 & 15.1 & 13.1 & 13.9 & 29.2 & 27.5  \\
        HT (w/ drop-key)     &  16.5 & 25.2 & 39.5 & 27.5 & 26.7 & 86.1 & 82.5  \\
        \bottomrule
        \end{tabular}
        \vspace{-10pt}
    \end{wraptable}
    
Environments (f) and (g) are more challenging for RL agents due to greater exit depth of their goal dependency graphs, \ie, the agent must achieve a longer sequence of intermediate goals before it can reach the exit.
Similarly, the width of the dependency graph introduces complexity (due to paths that don't lead anywhere), but not as much as the depth.
We find that the analytically estimated hitting times given in Table~\ref{tab:laplacian} are in agreement with the observed empirical performance of the RL algorithms.
We also note that despite the complexity of the environments, HiPPO is still able to make some progress on (f) and (g), while the other flat PPO baselines (with or without reward shaping and/or policy sketches) fail to solve them (Figure~\ref{fig:learning-plot}).

\def\varoffset{0.50}
\def\varlabel{-0.35}
\def\varstart{0}
\def\varone{\varstart+\varoffset}
\def\vartwo{\varstart+2*\varoffset}
\def\varthree{\varstart+3*\varoffset}
\def\varfour{\varstart+4*\varoffset}
\def\varfive{\varstart+5*\varoffset}
\def\varsix{\varstart+6*\varoffset}
\def\varseven{\varstart+7*\varoffset}

\def\imagewidth{80pt}
\begin{table}
        \caption{\textbf{Left}: Average success rate (\%) to reach the final goal over the last 10 training episodes. \textbf{Right}: Average episode length (\% of the max length, smaller is more efficient) over the last 10 training episodes. ``--'' indicates that the method failed to reach the final goal within 1000 steps.}
        \label{tab:success_rate}
        %\vspace{-1ex}
        \centering
        \scriptsize
        \setlength{\tabcolsep}{5.7pt}
        \begin{tabular}{@{}l|rrrrrrr|rrrrrrr@{}}
        \toprule
        & \multicolumn{7}{c}{Average Success Rate} & \multicolumn{7}{c}{Average Episode Length} \\
         & (a) & (b) & (c) & (d) & (e) & (f) & (g)
         & (a) & (b) & (c) & (d) & (e) & (f) & (g) \\
        \midrule
        PPO & 56.1 & 28.2 & 0.2 & 22.7 & 19.1 & 0.0 & 0.0
        & 78.5 & 88.0 & -- & 90.7 & 91.4 & -- & -- \\
        PPO+Bonus & 9.0 & 6.0 & 0.0 & 11.0 & 4.5 & 0.5 & 0.0
        & 97.8 & 96.7 & 99.9 & 97.3 & 98.0 & 99.9 & -- \\
        PPO+Sketch & 23.2 & 14.5 & 0.4 & 12.6 & 10.7 & 0.1 & 0.0 
        & 91.9 & 94.4 & 99.9 & 95.0 & 95.7 & -- & -- \\
        HiPPO & \textbf{74.9} & \textbf{57.0} & \textbf{60.8} & \textbf{48.0} & \textbf{29.9} & \textbf{11.2} & \textbf{19.0}
        & \textbf{48.2} & \textbf{67.4} & \textbf{69.1} & \textbf{71.0} & \textbf{85.3} & \textbf{96.4} & \textbf{93.8} \\
        \bottomrule
        \end{tabular}
\end{table}

\begin{wrapfigure}{r}{0.5\textwidth}
    \centering
    \includegraphics[width=0.5\textwidth]{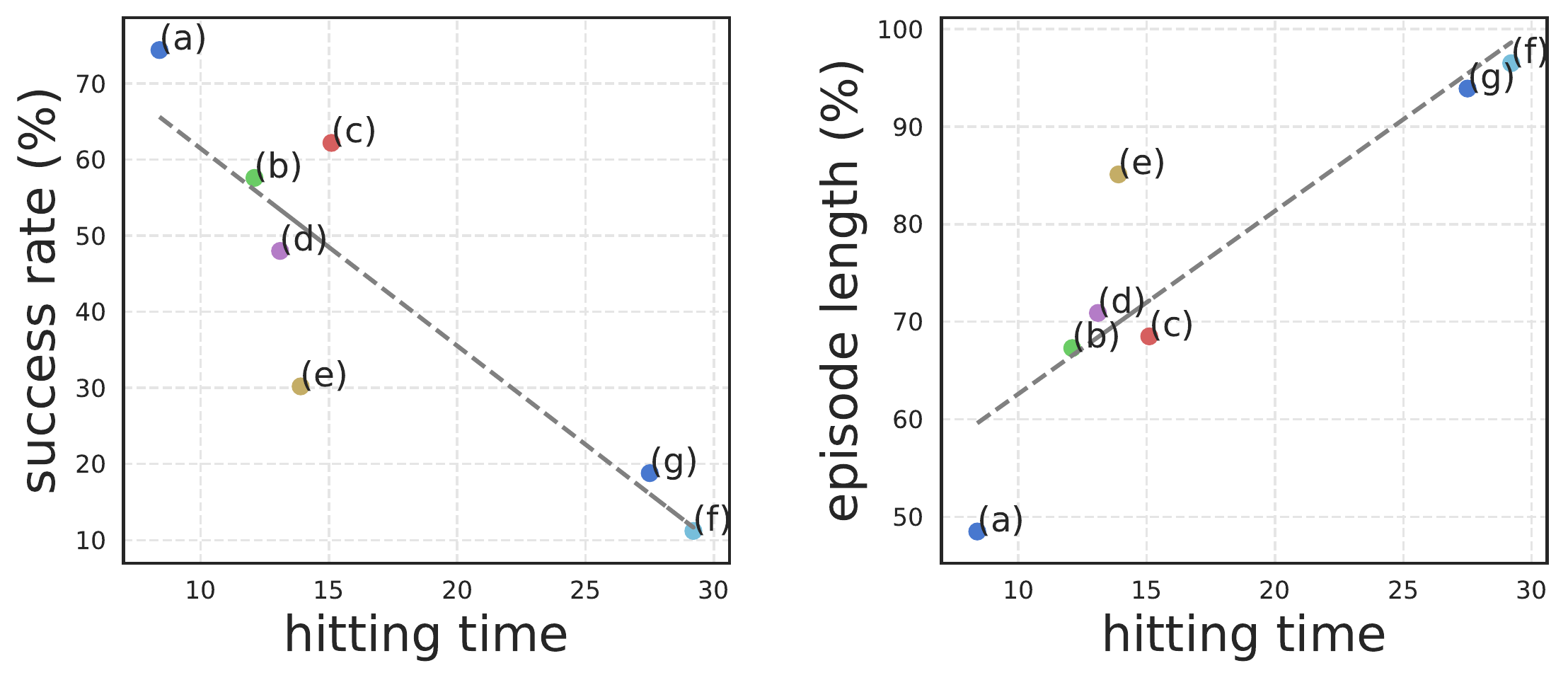}
    \caption{Correlation between the hitting time (Table~\ref{tab:laplacian}) vs. the average success rate and average episode length (Table~\ref{tab:success_rate}) of HiPPO for \texttt{EscapeRoom} environments (a)-(g) from Table~\ref{tab:env-escaperoom}. This verifies that the hitting time is a useful measure of complexity for RL environments.}
    \vspace{-30pt}
    \label{fig:ht-correlation}
\end{wrapfigure}

In Figure~\ref{fig:ht-correlation}, we illustrate the correlation between the hitting time on goal dependency graphs (Table~\ref{tab:laplacian}) and the empirical performance of HiPPO (Table~\ref{tab:success_rate}) for different \texttt{EscapeRoom} environments, which demonstrates that analytically estimated hitting time is an informative metric for measuring the complexity of an environment.
%!TEX root = ../main.tex

\section{Discussion}\label{sec:discussion}

We designed a simple grid-world \texttt{EscapeRoom} environment where it is easy to measure the exploration complexity by analyzing the corresponding goal dependency graphs.
We showed that hitting times in goal dependency graphs are consistent with the empirical performance of PPO-based methods, and is therefore a useful metric to measure the complexity of the environment.
Finally, we showed the performance improvement of HiPPO over other flat PPO baselines, demonstrating the benefit of using hierarchical policies that operate at different temporal scales.

\bibliography{references}

\begin{thebibliography}{15}
\providecommand{\natexlab}[1]{#1}
\providecommand{\url}[1]{\texttt{#1}}
\expandafter\ifx\csname urlstyle\endcsname\relax
  \providecommand{\doi}[1]{doi: #1}\else
  \providecommand{\doi}{doi: \begingroup \urlstyle{rm}\Url}\fi

\bibitem[Levine et~al.(2016)Levine, Finn, Darrell, and Abbeel]{levine2016end}
Sergey Levine, Chelsea Finn, Trevor Darrell, and Pieter Abbeel.
\newblock End-to-end training of deep visuomotor policies.
\newblock \emph{The Journal of Machine Learning Research}, 17\penalty0
  (1):\penalty0 1334--1373, 2016.

\bibitem[Mnih et~al.(2015)Mnih, Kavukcuoglu, Silver, Rusu, Veness, Bellemare,
  Graves, Riedmiller, Fidjeland, Ostrovski, et~al.]{mnih2015human}
Volodymyr Mnih, Koray Kavukcuoglu, David Silver, Andrei~A Rusu, Joel Veness,
  Marc~G Bellemare, Alex Graves, Martin Riedmiller, Andreas~K Fidjeland, Georg
  Ostrovski, et~al.
\newblock Human-level control through deep reinforcement learning.
\newblock \emph{Nature}, 518\penalty0 (7540):\penalty0 529--533, 2015.

\bibitem[Silver et~al.(2017)Silver, Schrittwieser, Simonyan, Antonoglou, Huang,
  Guez, Hubert, Baker, Lai, Bolton, et~al.]{silver2017mastering}
David Silver, Julian Schrittwieser, Karen Simonyan, Ioannis Antonoglou, Aja
  Huang, Arthur Guez, Thomas Hubert, Lucas Baker, Matthew Lai, Adrian Bolton,
  et~al.
\newblock Mastering the game of go without human knowledge.
\newblock \emph{Nature}, 550\penalty0 (7676):\penalty0 354, 2017.

\bibitem[Dann et~al.(2016)Dann, Hofmann, and Nowozin]{dann2016memory}
Christoph Dann, Katja Hofmann, and Sebastian Nowozin.
\newblock Memory lens: How much memory does an agent use?
\newblock \emph{arXiv preprint arXiv:1611.06928}, 2016.

\bibitem[Bellemare et~al.(2016)Bellemare, Srinivasan, Ostrovski, Schaul,
  Saxton, and Munos]{bellemare2016unifying}
Marc Bellemare, Sriram Srinivasan, Georg Ostrovski, Tom Schaul, David Saxton,
  and Remi Munos.
\newblock Unifying count-based exploration and intrinsic motivation.
\newblock In \emph{Advances in Neural Information Processing Systems}, pages
  1471--1479, 2016.

\bibitem[Fortunato et~al.(2017)Fortunato, Azar, Piot, Menick, Osband, Graves,
  Mnih, Munos, Hassabis, Pietquin, et~al.]{fortunato2017noisy}
Meire Fortunato, Mohammad~Gheshlaghi Azar, Bilal Piot, Jacob Menick, Ian
  Osband, Alex Graves, Vlad Mnih, Remi Munos, Demis Hassabis, Olivier Pietquin,
  et~al.
\newblock Noisy networks for exploration.
\newblock \emph{arXiv preprint arXiv:1706.10295}, 2017.

\bibitem[Plappert et~al.(2017)Plappert, Houthooft, Dhariwal, Sidor, Chen, Chen,
  Asfour, Abbeel, and Andrychowicz]{plappert2017parameter}
Matthias Plappert, Rein Houthooft, Prafulla Dhariwal, Szymon Sidor, Richard~Y
  Chen, Xi~Chen, Tamim Asfour, Pieter Abbeel, and Marcin Andrychowicz.
\newblock Parameter space noise for exploration.
\newblock \emph{arXiv preprint arXiv:1706.01905}, 2017.

\bibitem[Sutton et~al.(1999)Sutton, Precup, and Singh]{sutton1999between}
Richard~S Sutton, Doina Precup, and Satinder Singh.
\newblock Between mdps and semi-mdps: A framework for temporal abstraction in
  reinforcement learning.
\newblock \emph{Artificial intelligence}, 112\penalty0 (1-2):\penalty0
  181--211, 1999.

\bibitem[Bacon et~al.(2017)Bacon, Harb, and Precup]{bacon2017option}
Pierre-Luc Bacon, Jean Harb, and Doina Precup.
\newblock The option-critic architecture.
\newblock In \emph{AAAI}, pages 1726--1734, 2017.

\bibitem[Kulkarni et~al.(2016)Kulkarni, Narasimhan, Saeedi, and
  Tenenbaum]{kulkarni2016hierarchical}
Tejas~D Kulkarni, Karthik Narasimhan, Ardavan Saeedi, and Josh Tenenbaum.
\newblock Hierarchical deep reinforcement learning: Integrating temporal
  abstraction and intrinsic motivation.
\newblock In \emph{Advances in neural information processing systems}, pages
  3675--3683, 2016.

\bibitem[Schulman et~al.(2017)Schulman, Wolski, Dhariwal, Radford, and
  Klimov]{schulman2017proximal}
John Schulman, Filip Wolski, Prafulla Dhariwal, Alec Radford, and Oleg Klimov.
\newblock Proximal policy optimization algorithms.
\newblock \emph{arXiv preprint arXiv:1707.06347}, 2017.

\bibitem[Lov{\'a}sz(1993)]{lovasz1993random}
L{\'a}szl{\'o} Lov{\'a}sz.
\newblock Random walks on graphs.
\newblock \emph{Combinatorics, Paul erdos is eighty}, 2\penalty0
  (1-46):\penalty0 4, 1993.

\bibitem[Chevalier-Boisvert and Willems(2018)]{gym_minigrid}
Maxime Chevalier-Boisvert and Lucas Willems.
\newblock {Minimalistic Gridworld Environment for OpenAI Gym}.
\newblock \url{https://github.com/maximecb/gym-minigrid}, 2018.

\bibitem[Brockman et~al.(2016)Brockman, Cheung, Pettersson, Schneider,
  Schulman, Tang, and Zaremba]{brockman2016openai}
Greg Brockman, Vicki Cheung, Ludwig Pettersson, Jonas Schneider, John Schulman,
  Jie Tang, and Wojciech Zaremba.
\newblock Openai gym.
\newblock \emph{arXiv preprint arXiv:1606.01540}, 2016.

\bibitem[Andreas et~al.(2016)Andreas, Klein, and Levine]{andreas2016modular}
Jacob Andreas, Dan Klein, and Sergey Levine.
\newblock Modular multitask reinforcement learning with policy sketches.
\newblock \emph{arXiv preprint arXiv:1611.01796}, 2016.

\end{thebibliography}
\bibliographystyle{unsrtnat}
\newpage
\begin{appendix}
%!TEX root = ../main.tex

\section{Details on computing hitting times}
\label{app:laplacian}

As we mentioned in Section 2.3, to compute the hitting time of random walk we need an augmented goal-dependency graph (which can be generated procedurally from the graphs given in the main text).
An example augmented graph for \texttt{EscapeRoom} (c) from Table~\ref{tab:env-escaperoom} is presented below.

\begin{minipage}{\columnwidth}
    \centering
    \begin{tikzpicture}[->, >=stealth', shorten >=1pt, auto, node distance=1cm, thick, main/.style={circle,draw,minimum width=15pt}, x=1.5cm, y=3cm]
        % Nodes
        \node[main, fill=gray!30, draw=black!50, label={above:\scriptsize{start}}] (start) at (0, 0) {};
        \draw (start) to[out=0,in=45,loop,looseness=10] (start);
        
        \node[draw=none] (level1) at (2, 0) {All doors closed};
        
        \draw [-, dotted, draw=black!50] (-2.3, -0.2) -- (2.8, -0.2);
        
        \node[fill=blue!75, draw=black!50, label={left:\scriptsize{key 1}}] (key1) at (0, -0.4) {};
        \draw (key1) to[out=0,in=45,loop,looseness=10] (key1);
        
        \node[main, fill=blue!75, draw=black!50, label={-10:\scriptsize{room 1}}] (room1) at (0, -0.7) {};
        \draw (room1) to[out=0,in=45,loop,looseness=10] (room1);
        \node[main, fill=gray!30, draw=black!50, label={above:\scriptsize{start}}] (start1) at (-0.75, -0.7) {};
        \draw (start1) to[out=180,in=135,loop,looseness=10] (start1);
        
        \node[draw=none] (level3) at (2, -0.7) {First door open};
        
        \draw [-, dotted, draw=black!50] (-2.3, -0.9) -- (2.8, -0.9);
        
        \node[fill=purple!75, draw=black!50, label={left:\scriptsize{key 2}}] (key2) at (0, -1.1) {};
        \draw (key2) to[out=0,in=45,loop,looseness=10] (key2);
        
        \node[main, fill=purple!75, draw=black!50, label={-10:\scriptsize{room 2}}] (room2) at (0, -1.4) {};
        \draw (room2) to[out=0,in=45,loop,looseness=10] (room2);
        \node[main, fill=blue!75, draw=black!50, label={below:\scriptsize{room 1}}] (room12) at (-0.75, -1.4) {};
        \draw (room12) to[out=120,in=60,loop,looseness=7] (room12);
        \node[main, fill=gray!30, draw=black!50, label={above:\scriptsize{start}}] (start2) at (-1.5, -1.4) {};
        \draw (start2) to[out=180,in=135,loop,looseness=10] (start2);
        
        \node[draw=none] (level3) at (2, -1.4) {Both doors open};
        
        \draw [-, dotted, draw=black!50] (-2.3, -1.65) -- (2.8, -1.65);
        
        \node[main, fill=green!75, draw=black!50, label={below:\scriptsize{exit}}] (exit) at (0, -1.9) {};
        \draw (exit) to[out=180,in=135,loop,looseness=10] (exit);
        
        \node[draw=none] (level3) at (2, -1.9) {Found exit};
        
        % Edges
        \draw (start) to (key1);
        \draw (key1) to (room1);
        \draw (start1) to[out=30,in=150] (room1);
        \draw (room1) to[out=-140,in=-30] (start1);
        \draw (room1) to (key2);
        \draw (key2) to (room2);
        \draw (room12) to[out=30,in=150] (room2);
        \draw (room2) to[out=-140,in=-30] (room12);
        \draw (start2) to[out=30,in=150] (room12);
        \draw (room12) to[out=-140,in=-30] (start2);
        \draw (room2) to (exit);
    \end{tikzpicture}
\end{minipage}

The main difference from the goal dependency graph given in Table~\ref{tab:env-escaperoom} is that when the agent picks up a key and opens the corresponding door, it transitions into a subgraph that corresponds to the new layout of the rooms accessible to the agent.
Self-loops and transitions between the rooms represent the moving behavior.

We set the following parameters for the random walk.
With 80\% chance, no transition happens.
With 19\% chance, the walk transitions from the current node along one of the outgoing edges.
Finally, to ensure strong connectivity of the graph, we add 1\% chance of the agent moving back to the root start node from any other node in the graph\footnote{A similar approach is taken by the PageRank algorithm.}.
This corresponds to the situation where the agent is not able to reach the exit within the time limit and must start a new episode.
\end{appendix}

\end{document}